\documentclass[runningheads]{llncs}
\usepackage[T1]{fontenc}
\usepackage{graphicx,verbatim}
\usepackage{booktabs}
\usepackage{ulem}
\usepackage{float}
\usepackage{xcolor}
\usepackage{amsmath}

\begin{document}
\title{AffordTissue: Dense Affordance Prediction for Tool-Action Specific Tissue Interaction}
%

\author{Aiza Maksutova\inst{1,\thanks{Equal contribution.}}, Lalithkumar	Seenivasan\inst{1, \star}, Hao	Ding\inst{1, \star}, Jiru Xu\inst{1}, Chenhao	Yu\inst{1}, Chenyan	Jing\inst{1}, Yiqing Shen\inst{1}, Mathias Unberath\inst{1}}  
\authorrunning{Maksutova et al.}
\institute{Johns Hopkins University, Baltimore MD, USA \\
    \email{\{\text{unberath}\}\text{@jhu.edu}}}
\maketitle              
\begin{abstract}
Surgical action automation has progressed rapidly toward achieving surgeon-like dexterous control, driven primarily by advances in learning from demonstration and vision-language-action models. While these have demonstrated success in table-top experiments, translating them to clinical deployment remains challenging: current methods offer limited predictability on where instruments will interact on tissue surfaces and lack explicit conditioning inputs to enforce tool-action-specific safe interaction regions. Addressing this gap, we introduce AffordTissue, a multimodal framework for predicting tool-action specific tissue affordance regions as dense heatmaps during cholecystectomy. Our approach combines a temporal vision encoder capturing tool motion and tissue dynamics across multiple viewpoints, language conditioning enabling generalization across diverse instrument-action pairs, and a DiT-style decoder for dense affordance prediction. We establish the first tissue affordance benchmark by curating and annotating 15,638 video clips across 103 cholecystectomy procedures, covering six unique tool-action pairs involving four instruments (hook, grasper, scissors, clipper) and their associated tasks: dissection, grasping, clipping, and cutting. Experiments demonstrate substantial improvement over vision-language model baselines (20.6 px ASSD vs. 60.2 px for Molmo-VLM), showing that our task-specific architecture outperforms large-scale foundation models for dense surgical affordance prediction. By predicting tool-action specific tissue affordance regions, AffordTissue provides explicit spatial reasoning for safe surgical automation, potentially unlocking explicit policy guidance toward appropriate tissue regions and early safe stop when instruments deviate outside predicted safe zones.
\end{abstract}

\keywords{Affordance Prediction \and Tissue Affordance \and Surgical Automation  \and Surgical Scene Understanding \and Multimodal Models}

\section{Introduction}
Recent advances in learning from demonstration~\cite{chi2025diffusion}, imitation policies~\cite{zitkovich2023rt,team2024octo}, and vision-language-action models (VLAs)~\cite{kim2024openvla,black2024pi_0} have significantly accelerated surgical task automation, with pioneering works demonstrating surgeon-like dexterous control in controlled table-top experiments~\cite{kim2025srt,scheikl2022sim}. 
However, with safety taking precedence over efficiency in clinical practice, truly translating these advances to clinical deployment hinges on ensuring safe automation~\cite{attanasio2021autonomy,haidegger2019autonomy}.
A fundamental limitation of current learning-based approaches lies in their ``black box'' nature: while these models learn complex dexterous manipulation from expert demonstrations, they offer limited predictability regarding \textit{where} and \textit{how} the robot will interact with tissue, and \textit{whether} the action will succeed~\cite{attanasio2021autonomy,haidegger2019autonomy}. Furthermore, they provide no explicit mechanism to condition or verify safe interaction. This poses significant safety concerns -- 
there is limited controllability in enforcing action-specific tissue interaction zones or intervening before potentially harmful contact occurs.

Current surgical scene understanding approaches rely predominantly on semantic segmentation~\cite{maier2022surgical,allan20202018}, which can identify anatomical structures but lacks \textit{interaction-aware} perception-reasoning about \textit{where within} those structures a specific tool can safely engage for a given action. Affordance prediction has emerged as a powerful paradigm for interaction-aware perception in robotic manipulation, enabling systems to identify and enforce \textit{where} and \textit{how} objects can be acted upon~\cite{xu2025a0,nagarajan2019grounded}. However, these methods predominantly target rigid-object interaction, with limited exploration of deformable tissue dynamics, context-dependent surgical constraints, and safety-critical medical requirements. 

To this end, we introduce \textbf{AffordTissue}, a multimodal framework for predicting tissue affordance regions as dense heatmaps conditioned on tool-action specifications. Similar to costmaps in robot navigation~\cite{lu2014layered}, our affordance representation encodes interaction suitability with maximum affordance at the region center that decreases toward boundaries. This potentially unlocks two complementary modes for safe automation: (i) a conditioning signal, guiding learned policies toward appropriate tissue regions, and (ii) a safety layer, triggering automated early robot stopping when instrument trajectories deviate toward tissue outside the predicted affordance region -- \textit{before} potentially harmful contact occurs. 
%
\textbf{Our key contributions are}: (i) we introduce dense tissue affordance prediction as a novel task that provides explicit spatial reasoning about tool-action specific affordance regions on tissue surfaces; (ii) we propose a multimodal architecture producing tool-action conditioned affordance heatmaps toward action guidance and safety verification; and (iii) we curate and annotate 103 cholecystectomy videos, establishing the first tissue affordance benchmark.

\section{AffordTissue}

Our proposed AffordTissue framework for predicting tool-action specific affordance reformulates image diffusion into a novel dense heatmap prediction task (Fig.~\ref{fig:architecture_workflow}). It takes in two inputs: (i) a language prompt specifying the tool–action pair, the surgical context, and the prediction objective, and (ii) a video sequence consisting of the target ($t_0$) and past frames ($t_{-256}\rightarrow t_{-1}$). The architecture includes (a) a language encoder that embeds the text prompt, (b) a temporal video transformer that encodes spatiotemporal visual information, and (c) a multimodal diffusion decoder that fuses the both to produce a task-aware dense tool-tissue interaction heatmap for the target frame. We begin by detailing the pre-trained backbones and architectural components integrated into our pipeline. Afterwards, we formalize the input space and the end-to-end workflow.

\subsection{Preliminaries}
\noindent\textbf{(a) SigLIP 2~\cite{tschannen2025siglip}}: SigLIP 2 is built on SigLIP~\cite{zhai2023sigmoid}, a CLIP-style model that employs a \textbf{pairwise sigmoid loss} instead of a conventional softmax loss used in CLIP, reducing the model's dependency on batch normalization and improving training efficiency. SigLIP 2 further incorporates captioning-based pretraining and self-supervised losses, which significantly improve dense feature representation for segmentation and localization.

\noindent\textbf{(b) Video Swin Transformer~\cite{liu2022video}:} Video Swin Transformer is a hierarchical backbone that extends the 2D Swin Transformer architecture to the spatiotemporal domain. The model processes an input video by partitioning it into 3D tokens, treating each 3D patch as a unique token. It employs a \textit{3D Shifted Window} mechanism to partition the tokens into non-overlapping spatiotemporal windows, where self-attention is computed locally. To ensure cross-window and cross-frame information exchange, the window partitions are shifted along the T, H, and W axes in successive layers, allowing the model to attend to temporal dependencies between several frames.

\noindent\textbf{(c) Adaptive Layer Normalization (AdaLN) Decoder}: 
Introduced in DiT ~\cite{peebles2023scalable}, the decoder is designed to inject conditional information directly into the feature normalization layers of a transformer decoder. Unlike standard cross-attention, which computes pairwise similarities between all tokens, AdaLN performs global conditioning by regressing the scale $\gamma$ and shift $\beta$ parameters of the Layer Normalization from the input condition. By shifting and scaling activations based on the conditioning task, AdaLN enables efficient embedding fusion that emphasizes spatial regions relevant to the prediction objective, achieving computational efficiency while maintaining representational power.

\subsection{Pipeline}

\begin{figure}[!h]
    \centering
    \includegraphics[width=0.91\textwidth]{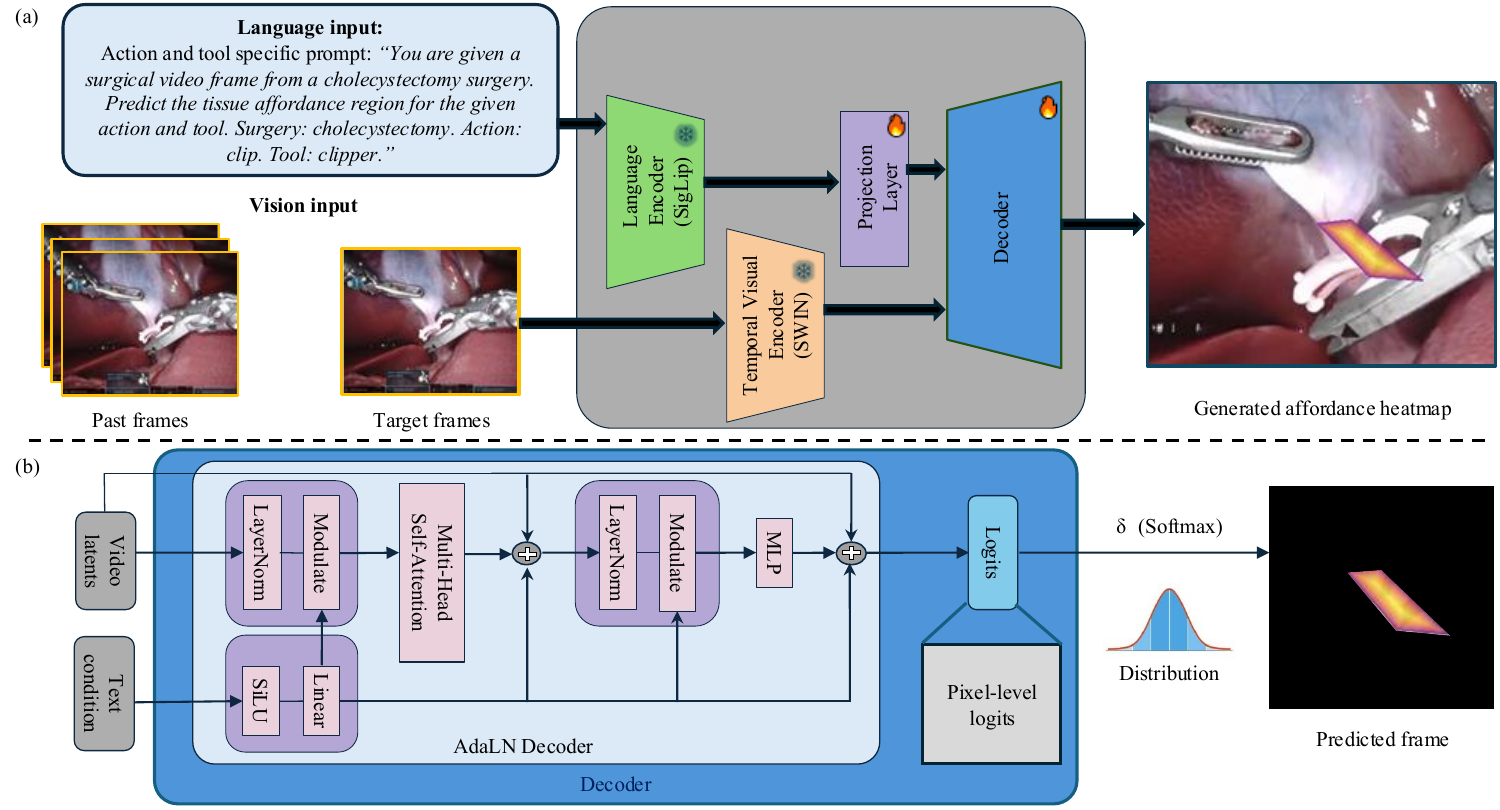}
    \caption{Architecture of AffordTissue: a) High-level overview: The pipeline processes tool-action prompts and temporal vision context via frozen SigLIP and Swin encoders, with a trained AdaLN-based decoder fusing these embeddings to output a heatmap. b) Decoder details: The decoder utilizes text-conditioned adaptive layer normalization (AdaLN) to refine temporal vision latents for a specific tool-action pair. This results in pixel-level logits that estimate the probability of each pixel belonging to the safe tissue affordance region.}
    \label{fig:architecture_workflow}
\end{figure}


\noindent\textbf{(a) Text input:} Incorporating text in the pipeline is a crucial step for model scalability, as it allows it to differentiate between different tools and actions during the same surgery. We found that simple prompts work best for such tasks. Our chosen prompt consists of a surgical triplet: \textit{\{surgery type, tool type, action type\}}. From VLM best practices, we also define the model's final objective inside the prompt. 

\noindent\textbf{(b) Visual input:} We input a temporal window of $N=256$ frames (stride of 8) to capture tool motion and tissue dynamics from multiple viewpoints, enabling implicit modeling of deformation patterns and surgical intent beyond single-frame approaches.


\noindent\textbf{(c) Decoder adaptation:} We propose a task-specific adaptation of the AdaLN decoder for dense heatmap prediction, detailed in Fig.~\ref{fig:architecture_workflow}(b). While DiT employs AdaLN to predict diffusion noise from noised latent representations conditioned on class labels, our architecture replaces both the input space and the objective. Our decoder directly processes temporal vision embeddings with language embeddings as condition to predict per-pixel logits, estimating the probability that each pixel belongs to the tissue affordance region. This shift demonstrates that AdaLN-based conditioning, originally designed for diffusion modeling, can be effectively repurposed for spatially grounded dense feature prediction in vision-language tasks.

\noindent\textbf{(d) Workflow:} Our end-to-end pipeline is shown in Fig.~\ref{fig:architecture_workflow}(a). Given a text prompt, the SigLIP 2 encoder produces a $(B, 1152)$ embedding, which is afterwards projected to a shared embedding space using MLP. Concurrently, the Video Swin Transformer processes input frames of shape $(B, C, T, H, W)$ to extract spatiotemporal features of shape $(B, C, H, W)$. The AdaLN decoder then fuses these representations to predict per-pixel logits, producing a probabilistic affordance heatmap.

\section{Experiment}

\begin{table}[!t]
    \centering
    \small 
    \setlength{\tabcolsep}{5pt} 
    \caption{Distribution of video clips per tool–action pair across datasets. Each entry describes the total number of clips, with (train/val/test) split shown in parentheses.}
    \label{tab:tool_action_distribution}
    \scalebox{0.65}{
    \begin{tabular}{lccccc}
        \toprule
        & \multicolumn{5}{c}{\textbf{Dataset type}} \\
        \textbf{Tool-action pair} & \textbf{Youtube} & \textbf{Cholec80} & \textbf{HeiChole} & \textbf{CHEC} & \textbf{SurgVU} \\
        & (21 videos) & (34 videos) & (11 videos) & (8 videos) & (29 videos) \\
        \midrule
        Dissect - Hook 
        & 1280 (1066/41/173) 
        & 4726 (3454/741/531) 
        & 1106 (875/97/134) 
        & 628 (476/0/152) 
        & 2685 (2120/270/295) \\

        Dissect - Grasper 
        & 493 (462/0/31) 
        & 43 (36/0/7) 
        & 101 (92/5/4) 
        & 0 
        & 295 (234/61/1) \\

        Dissect - Scissors 
        & 209 (209/0/0) 
        & 12 (12/0/0) 
        & 100 (97/3/0) 
        & 0 
        & 350 (350/0/0) \\

        Grasp - Grasper 
        & 584 (538/3/43) 
        & 931 (677/189/65) 
        & 321 (286/9/26) 
        & 136 (97/0/39) 
        & 799 (689/66/44) \\

        Clip - Clipper  
        & 125 (111/4/10) 
        & 185 (143/26/16) 
        & 87 (73/6/8) 
        & 0 
        & 99 (76/14/9) \\

        Cut - Scissors 
        & 69 (63/0/6) 
        & 145 (106/20/19) 
        & 79 (70/4/5) 
        & 0 
        & 49 (46/3/0) \\ 
        \bottomrule
    \end{tabular}}
\end{table}

\label{subsec:experimental_setup}
\textbf{(i) Dataset:} We curate a custom dataset of 15638 video clips from 103 videos: Youtube (21 videos), Cholec-80 (34 videos)~\cite{twinanda2016endonet}, HeiChole (11 videos) \cite{wagner2023comparative}, Comprehensive Robotic Cholecystectomy Dataset (8 videos)~\cite{oh2024expanded}, and  SurgVU (29 videos)~\cite{zia2025surgical}. Table~\ref{tab:tool_action_distribution} details the distribution of video clips across datasets and train/validation/test splits. Each video clip is annotated for tool-action pairs (language) and tissue affordance. To define target affordance regions, each case was manually annotated with four keypoints outlining the safe tool–tissue interaction zone. These keypoints form a polygon, from which a target heatmap is generated by centering a Gaussian distribution at the polygon’s centroid. Since the objective is to predict tissue affordance prior to instrument interaction, only frames occurring before the onset of the surgical action are used. The dataset split is performed at the case level, ensuring no data leakage. 

\noindent \textbf{(ii) Training and inference:}  In our model, the language and the vision encoder are frozen during training and used solely to provide embeddings, and the decoder parameters are optimized. The model is trained on a single NVIDIA A100 GPU for 100 epochs, using a combination of binary cross-entropy loss and soft intersection-over-union (IoU) loss -- a differentiable approximation of the standard IoU metric. Optimization is performed using AdamW~\cite{loshchilov2017decoupled} with an initial learning rate of $1 \times 10^{-4}$ and a cosine learning rate scheduler. During training, we select random target frames within the pre-action range. Along with each target frame, we input 256 previous frames with a stride of 8. This corresponds to approximately 10.6 seconds of historical context, which we found sufficient for capturing relevant temporal dynamics.

\noindent \textbf{(iii) Evaluation Metrics:} We divide our evaluation metrics into two groups: logits-conditioned and boundary-conditioned metrics. \uline{Logits-conditioned metrics} assess probabilistic overlap of target and predicted heatmaps and include "soft" Dice score~\cite{milletari2016v} computed directly on logits, corresponding to heatmap intensities for each pixel. \uline{Boundary-conditioned metrics} evaluate spatial alignment, and include Precision at K (PCK)~\cite{andriluka14cvpr}, Hausdorff Distance (HD) in pixels, and Average Symmetric Surface Distance (ASSD) in pixels~\cite{gerig2001valmet}. For each case, we evaluate eight pre-action frames, chosen with a preference to earlier timestamps. This reduces the likelihood that the model grounds its prediction on the instrument's position, which becomes closer to target tissue affordance area as the action approaches.

\begin{table*}[!b]
\centering
\small
\caption{Comparison of our model against baselines: Molmo-VLM~\cite{deitke2024molmo}, SAM3~\cite{sam3_2026}, and Qwen-VLM~\cite{qwen25_2024}.}
\scalebox{0.7}{
\begin{tabular}{lcccccc}
\toprule
\textbf{Metrics} &
\textbf{DICE$\uparrow$} &
\textbf{PCK@0.05$\uparrow$} &
\textbf{PCK@0.1$\uparrow$} &
\textbf{HD(px)$\downarrow$} &
\textbf{ASSD(px)$\downarrow$} \\

\midrule
\textbf{Ours} & 0.124 & \textbf{0.517} & \textbf{0.667} & \textbf{79.763} & \textbf{20.557}  \\
SAM3~\cite{sam3_2026} & \textbf{0.18} & 0.128 & 0.221 & 150.320 & 81.138 &  \\
Molmo-VLM~\cite{deitke2024molmo}  & 0.026 & 0.095 & 0.320 & 129.494 & 60.184 \\
Qwen-VLM (8B)~\cite{qwen25_2024}   & 0.014 & 0.031 & 0.022 & 203.214 & 111.271 \\
\bottomrule
\end{tabular}}
\label{tab:Baseline_Comparision}
\end{table*}

\section{Results and Ablation Study}
\begin{figure}[!t]
    \centering
    \includegraphics[width=1\textwidth]{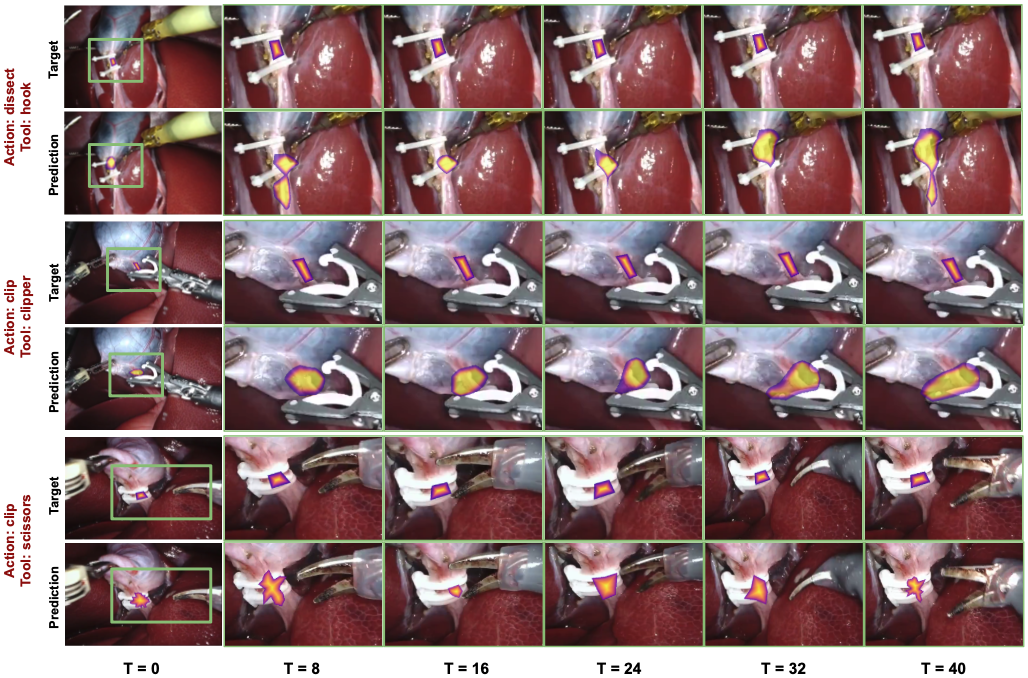}
    \caption{Qualitative comparison between ground truth and predicted heatmaps for three tool–action pairs: (i) hook - dissection, (ii) clipper - clipping, and (iii) scissors -clipping, across six representative timestamps.}
    \label{fig:results1}
\end{figure}

\textbf{(i) Baseline Comparison:} 
We quantitatively evaluate our pipeline's performance in detecting safe tool-tissue interaction against the following baseline models: Molmo-VLM~\cite{deitke2024molmo}, QWEN-8B VLM~\cite{qwen25_2024} and SAM3~\cite{sam3_2026}.
Due to the novelty of the proposed task, there are no publicly available baselines specifically trained for tissue affordance prediction. To provide meaningful comparison, we evaluate against closely related tasks, such as pointing and segmentation, which approximate different aspects of spatial affordance estimation. For the pointing baseline, we fine-tune Molmo-VLM and QWEN-8B VLM. We augment the architecture with additional regression heads to predict the four polygon vertices that define the tissue affordance region. For the segmentation baseline, we fine-tune SAM3. None of the above baselines produce a heatmap output directly. To enable a fair comparison, we convert their predictions into heatmap representations by generating a Gaussian distribution centered at the centroid of the predicted polygon or segmentation mask.
From Table~\ref{tab:Baseline_Comparision} we observe that the model achieves an \textbf{ASSD of 20.557 px} and \textbf{PCK@0.05 and PCK@0.1 of 0.517 and 0.667}, respectively. This strong spatial alignment between predicted and ground-truth heatmaps observed quantitatively is supported by a qualitative analysis in Fig.~\ref{fig:results1}. In contrast, the Hausforff Distance (HD) performance is relatively high. Qualitative analysis showed that it is primarily caused by occasional secondary, low-confidence heatmap activations that appear on the tool surface. We suggest that in future studies we add a carefully designed maximum-distance penalty to solve this problem. The DICE score is comparatively low across cases. As illustrated in Fig.~\ref{fig:results1}, this is largely due to differences in intensity distributions rather than boundary misalignment. Given that the affordance regions are annotated as uniformly safe for interaction, the intensity discrepancy does not pose a problem as of now. The standard IoU metric for localization is not added to our evaluation experiments, as it is included in our training loss.
Our model substantially outperforms all evaluated baselines. The strongest competitor, fine-tuned \textbf{Molmo-VLM,} \textbf{shows a degradation of 192.76\% in ASSD and 62.34\% in HD} relative to our approach. We also observe that SAM3 achieves a higher Dice score than our model. However, qualitative inspection of cases where SAM3 outperforms our approach reveals that it often predicts nearly the entire tissue region in the frame as safe for interaction, leading to higher DICE score but lower overall performance. These results indicate that even large-scale foundation models do not match the performance of our task-specific architecture for dense heatmap prediction in laparoscopic data.

\begin{table*}[!b]
\centering
\caption{Ablation study of our model with a different (i) language encoder, (ii) vision encoder, and (iii) decoder.}

\scalebox{0.7}{
\begin{tabular}{lcccccc}
\toprule
\textbf{Metrics} &
\textbf{DICE$\uparrow$} &
\textbf{PCK@0.05$\uparrow$} &
\textbf{PCK@0.1$\uparrow$} &
\textbf{HD(px)$\downarrow$} &
\textbf{ASSD(px)$\downarrow$} \\
\midrule
\textbf{Ours}       &\textbf{ 0.124} & \textbf{0.517} & \textbf{0.667} & \textbf{79.763} & \textbf{20.557} \\
{Ours with Bert-based language encoder}  & 0.106 & 0.502 & 0.651 & 84.403 & 23.891 \\
Ours with vision encoder & 0.086 & 0.422 & 0.609 & 113.613 & 24.329 &  \\
{Ours with cross-attention decoder}  & 0.085 & 0.443 & 0.571 & 108.039 & 28.736 \\

\bottomrule
\end{tabular}}

\label{tab:ABLATION_RESULTS_ARCHITECTURE_different}
\end{table*}

\begin{table*}[!b]
\centering
\caption{Ablation study of our model without (i) L (SigLip language encoder) and (ii) A (image augmentations).}
\scalebox{0.7}{
\begin{tabular}{lrrrrrr}
\toprule
\textbf{Metrics} &
\textbf{DICE$\uparrow$} &
\textbf{PCK@0.05$\uparrow$} &
\textbf{PCK@0.1$\uparrow$} &
\textbf{HD(px)$\downarrow$} &
\textbf{ASSD(px)$\downarrow$} \\
\midrule
\textbf{Ours}         &\textbf{ 0.124} & \textbf{0.517} & \textbf{0.667} & \textbf{79.763} & \textbf{20.557} \\
Ablation (w/o A)  & 0.094 & 0.491 & 0.625 & 127.101 & 29.181 \\
Ablation (w/o L)   & 0.068 & 0.205 & 0.348 & 170.482 & 43.135 \\
\bottomrule
\end{tabular}}


\label{tab:ABLATION_RESULTS_ARCHITECTURE_removal}
\end{table*}

\begin{table*}[!t]
\centering
\caption{Ablation study of our model without (i) previous frames, (ii) action, and (iii) tool specification in input.}
\scalebox{0.7}{
\begin{tabular}{lrrrrrr}
\toprule
\textbf{Metrics} &
\textbf{DICE$\uparrow$} &
\textbf{PCK@0.05$\uparrow$} &
\textbf{PCK@0.1$\uparrow$} &
\textbf{HD(px)$\downarrow$} &
\textbf{ASSD(px)$\downarrow$} \\

\midrule
\textbf{Ours}         &\textbf{ 0.124} & \textbf{0.517} & \textbf{0.667} & \textbf{79.763} & \textbf{20.557} \\
Ablation (w/o action)  & 0.112 & 0.350 & 0.542 & 104.733 & 22.087 \\
Ablation (w/o previous frames)   & 0.103 & 0.490 & 0.635 & 85.932 & 24.973 \\
Ablation (w/o tool)   & 0.101 & 0.320 & 0.495 & 93.702 & 27.302 \\
\bottomrule
\end{tabular}}

\label{tab:ABLATION_RESULTS_INPUT}
\end{table*}



\noindent \textbf{(ii) Ablation study:} We perform extensive ablation study to validate choice of every architecture and input component in the suggested workflow. Table~\ref{tab:ABLATION_RESULTS_ARCHITECTURE_different} demonstrates the impact of the SigLIP language encoder, the Swin vision encoder, and the AdaLN decoder. Among given modules, the decoder and temporal vision encoder choices have the most crucial effect on model's performance. Replacing AdaLN decoder with cross-attention decoder~\cite{vaswani2017attention} not only reduces memory efficiency, but also leads to a 35.45\% increase in HD and 39.78\% increase in ASSD. This highlights the importance of adaptive feature modulation in our setting, suggesting that global conditional normalization is better suited for structured heatmap prediction than token-level cross-attention fusion. Similarly, changing the Swin Transformer with a 3D ResNet-18~\cite{hara20183dresnet} leads to a significant performance degradation - (HD +42.43\%, ASSD +18.34\%), justifying our choice of vision backbone. Table~\ref{tab:ABLATION_RESULTS_ARCHITECTURE_removal} presents the importance of language encoder and dataset augmentations. The importance of language encoding is evident as the model's performance drops drastically when it is removed: \textbf{ASSD increases by 109.8\% and Hausdorff Distance by 213.7\%}. Given that the training set contains different tool–action pairs, the model must be given a mechanism to differentiate between them, making the language encoder a crucial part of the pipeline. In contrast, image augmentations provide a modest but consistent improvement, showing that while they contribute to robustness, they are secondary to model performance.

Table ~\ref{tab:ABLATION_RESULTS_INPUT} provides an ablation study for different input structures that we use in our pipeline: temporal vision context and action and tool specification in the prompt. Removing the tool specification leads to a substantial performance drop, with ASSD increasing by 32.81\%. This result is expected, since many samples in our dataset include frames with multiple instruments, making it confusing for the model to understand for which tool the affordance region should be predicted.
In comparison, removing temporal context is shown to be not as important to model's performance, leading to a more moderate decline (ASSD +21.48\%, HD +7.73\%). While less critical than text conditioning, temporal information still improves spatial consistency by giving the model multi-view context that refines tissue affordance prediction.

\section{Discussion and Conclusion}
We present AffordTissue, a multimodal framework combining a temporal vision encoder, language conditioning, and a DiT-style decoder for predicting tool-action specific tissue affordance regions as dense heatmaps for six unique tool-action pairs critical to during the cholecystectomy procedure. By explicitly conditioning on tool-action specifications and temporal context, our approach achieves more precise affordance localization than large foundational models such as SAM3 and Molmo-VLM. The predicted affordance maps potentially unlock two complementary modes for safe surgical automation: a conditioning signal guiding learned policies toward appropriate tissue regions, and a safety layer enabling early automated stopping when instruments deviate outside predicted affordance regions. Future directions include extending affordance prediction to ground on the surgical phase for stage-specific reasoning, expanding to model diverse tool-tissue actions, and integration with VLAs for closed-loop surgical automation.

\bibliographystyle{splncs04}
\bibliography{references}
\end{document}